\documentclass{sig-alternate-05-2015}
\newcommand{\tool}{{\tt graph2vec}}
\newcommand{\soa}{state-of-the-art}
\newcommand{\wv}{{\tt word2vec}}
\newcommand{\dv}{{\tt doc2vec}}
\newcommand{\nv}{{\tt node2vec}}
\newcommand{\sv}{{\tt sub2vec}}
\usepackage{amsmath,amssymb,amsfonts}
\usepackage{multirow}
\usepackage[table,xcdraw]{xcolor}
\usepackage[utf8]{inputenc}
\usepackage{cleveref}
\crefname{section}{§}{§§}
\Crefname{section}{§}{§§}
\usepackage{float}
\usepackage{graphicx}
\usepackage{hyperref}
\usepackage{color}
\hypersetup{
	colorlinks=true,
	linkcolor=blue,
	citecolor=blue,
	filecolor=black,
	urlcolor=blue,
}

\makeatletter
\DeclareUrlCommand\ULurl@@{%
	\def{\scriptsize ($ \pm 0.00 $)}Font{\ttfamily\color{blue}}%
	\def{\scriptsize ($ \pm 0.00 $)}Left{\uline\bgroup}%
	\def{\scriptsize ($ \pm 0.00 $)}Right{\egroup}}
\def\ULurl@#1{\hyper@linkurl{\ULurl@@{#1}}{#1}}
\DeclareRobustCommand*\ULurl{\hyper@normalise\ULurl@}
\makeatother

\usepackage{array}
\usepackage{pifont}
\newcolumntype{L}[1]{>{\raggedright\let\newline\\\arraybackslash\hspace{0pt}}m{#1}}
\newcolumntype{C}[1]{>{\centering\let\newline\\\arraybackslash\hspace{0pt}}m{#1}}
\newcolumntype{R}[1]{>{\raggedleft\let\newline\\\arraybackslash\hspace{0pt}}m{#1}}

\usepackage[font=small,labelfont=bf,
justification=justified,
format=plain]{caption}
\DeclareCaptionType{copyrightbox}
\usepackage{amsfonts}
\usepackage{fixltx2e}
\usepackage{bm}
\usepackage{amsmath}
\usepackage{graphicx}
\usepackage{amssymb}
\usepackage{tikz}
\usepackage{array}
\usepackage{xcolor}
\usepackage[mathscr]{euscript}
\usepackage{mathrsfs}
\usepackage{tikz}

\usetikzlibrary{shapes.geometric, arrows}
\tikzstyle{ADG} = [ellipse, minimum width=1cm, minimum height=.75cm,text centered, draw=black, fill=pink!30]
\tikzstyle{PDG} = [ellipse, minimum width=1cm, minimum height=.75cm,text centered, draw=black, fill=yellow!20]
\tikzstyle{SSCFP} = [ellipse, minimum width=1cm, minimum height=.75cm,text centered, draw=black, fill=green!10]
\tikzstyle{Ins} = [ellipse, minimum width=1cm, minimum height=.75cm,text centered, draw=black, fill=brown!10]
\tikzstyle{Signs} = [ellipse, minimum width=1cm, minimum height=.75cm,text centered, draw=black, fill=orange!10]

\usepackage{listings}
\lstdefinestyle{customc}{
	belowcaptionskip=1\baselineskip,
	breaklines=true,
	frame=L,
	xleftmargin=\parindent,
	language=Java,
	showstringspaces=false,
	basicstyle=\footnotesize\ttfamily,
	keywordstyle=\bfseries\color{green!40!black},
	commentstyle=\itshape\color{purple!40!black},
	identifierstyle=\color{blue},
	stringstyle=\color{orange},
}

\usepackage{pifont}
\usepackage[ruled,vlined,linesnumbered]{algorithm2e}

\usepackage{verbatim}

\usepackage{enumitem}
\usepackage[space]{cite}
\usepackage{footmisc}
\usepackage{footnote}
\makeatletter
\newtoks\therules
\therules={}
\def\appendto#1#2{\expandafter#1\expandafter{\the#1#2}}
\def\gobblefirst#1{
	#1\expandafter\expandafter\expandafter{\expandafter\@gobble\the#1}}%
\def\LState{\State\unskip\the\therules}
\def\printindent{\unskip\the\therules}%

\begin{document}
\sloppy

\title{graph2vec: Learning Distributed Representations of Graphs}
\numberofauthors{1}
\author{
	\alignauthor
	Annamalai Narayanan, Mahinthan Chandramohan, Rajasekar Venkatesan, Lihui Chen, Yang Liu and Shantanu Jaiswal\\
	\affaddr{Nanyang Technological University, Singapore}\\
	\email{\texttt{annamala002@e.ntu.edu.sg, \{mahinthan,rajasekarv,elhchen,yangliu\}@ntu.edu.sg,shantanu004@e.ntu.edu.sg}}\\
}
\maketitle
 
\begin{abstract}
Recent works on representation learning for graph structured data predominantly focus on learning distributed representations of graph substructures such as nodes and subgraphs. However, many graph analytics tasks such as graph classification and clustering require representing entire graphs as fixed length feature vectors. While the aforementioned approaches are naturally unequipped to learn such representations, graph kernels remain as the most effective way of obtaining them. However, these graph kernels use handcrafted features (e.g., shortest paths, graphlets, etc.) and hence are hampered by problems such as poor generalization.
To address this limitation, in this work, we propose a neural embedding framework named \tool{} to learn data-driven distributed representations of arbitrary sized graphs. \tool's embeddings are learnt in an unsupervised manner and are task agnostic. Hence, they could be used for any downstream task such as graph classification, clustering and even seeding supervised representation learning approaches. Our experiments on several benchmark and large real-world datasets show that \tool{} achieves significant improvements in classification and clustering accuracies over substructure representation learning approaches and are competitive with state-of-the-art graph kernels.
\end{abstract}

%
%
%

%
%

%
%
\printccsdesc


\keywords{Graph Kernels, Deep Learning, Representation Learning

\section{Introduction}
\label{sec:intro}

Graph-structured data are ubiquitous nowadays in many domains such as social networks, cybersecurity, bio- and chemo-informatics. Many analytics tasks in these domains such as graph classification, clustering and regression require representing graphs as fixed-length feature vectors to facilitate applying appropriate Machine Learning (ML) algorithms. For instance, vectorial representations (\textit{aka} embeddings) of programs’ call graphs could be used to detect malware \cite{sg2vec} and those of chemical compounds could be used to predict their properties such as solubility and anti-cancer activity \cite{dgk}.

\textbf{Graph Kernels and handcrafted features.} Graph kernels are one of the most prominent ways of catering the aforementioned graph analytics tasks. Graph kernels evaluate the similarity (\textit{aka} kernel value) between a pair of graphs G and G’ by recursively decomposing them into atomic substructures (e.g., random walks, shortest paths, graphlets etc.) and defining a similarity (\textit{aka} kernel) function over the substructures (e.g., counting the number of common substructures across G and G’). Subsequently, kernel methods (e.g., Support Vector Machines (SVMs)) could be used for performing classification/clustering. However, these kernels exhibit two critical limitations: (1) Many of them do not provide explicit graph embeddings.
This renders using general purpose ML algorithms which operate on vector embeddings (e.g., Random Forests (RFs), Neural Networks, etc.) unusable with graph data.  (2) The substructures (i.e., walk, paths, etc.) leveraged by these kernels are ‘handcrafted’ i.e., determined manually with specific well-defined functions that help extracting such substructures from graphs. 
However, as noted by Yanardag and Vishwanathan \cite{dgk}, when such handcrafted features are used on large datasets of graphs, it leads to building very high dimensional, sparse and non-smooth representations and thus yield poor generalization. We note that replacing handcrafted features with ones that are learnt automatically from data could help to fix both the aforementioned limitations. In fact, in related domains such as text mining and computer vision, feature learning based approaches have outperformed handcrafted ones significantly across many tasks \cite{w2v,patchy}.

\textbf{Learning substructure embeddings.}
Recently, several approaches have been proposed to learn embeddings of graph substructures such as nodes \cite{n2v}, paths \cite{dgk} and subgraphs \cite{sg2vec,sub2vec}. These embeddings can then be used directly in substructure based analytics tasks such as node classification, community detection and link prediction. However, these substructure representation learning approaches are incapable of learning representations of entire graphs and hence cannot be used for tasks such as graph classification. 
As we show through our experiments in \S \ref{sec:eval}, obtaining graph embeddings through trivial extensions such as averaging or max pooling over substructure embeddings leads to suboptimal results.  

\textbf{Learning task-specific graph embeddings.}
On the other hand, a few supervised approaches (i.e., ones that require class labels of graphs) to learn embeddings of entire graphs such as \textsc{Patchy-san} \cite{patchy} have been proposed very recently.
While they offer excellent performances in supervised learning tasks (e.g., graph classification) they pose two critical limitations that reduce their usability:  (1) Being deep neural network based representation learning approaches, they require large volume of labeled data to learn meaningful representations. Obviously, obtaining such datasets is a challenge in itself as it requires mammoth labeling effort. (2) The representations thus learnt are specific to one particular ML task and cannot be used or transferred to other tasks or problems. For instance, the graph embeddings for the chemical compounds in the MUTAG dataset (see \cite{dgk} for details) learnt using \cite{patchy} are specifically designed to predict whether or not a compound has  mutagenic effect on a bacterium. Hence, the same embeddings could not be used for any other tasks such as predicting the properties of the compounds. 
To circumvent these limitations, similar to the above mentioned substructure representation learning approaches, we need a completely unsupervised approach that can succinctly capture the generic characteristics of entire graphs in the form of their embeddings. To the best of our knowledge, there are no such techniques available. Hence driven by this motivation, in this work, we take the first steps towards learning task-agnostic representations of arbitrary sized graphs in an unsupervised fashion. 

\textbf{Our approach.} To this end, we propose and develop a neural embedding framework named \tool. Inspired by the success of  recently proposed neural document embedding models,  we extend the same to learn graph embeddings. These document embedding models exploit the way how words/word sequences compose documents to learn their embeddings. 
Analogically, in \tool, we propose to view an entire graph as a document and the rooted subgraphs around every node in the graph as words that compose the document and extend document embedding neural networks to learn representations of entire graphs.

To the best of our knowledge, \tool{} is the first neural embedding approach that learns representations of whole graphs and it offers the following key advantages over graph kernels and other substructure embedding approaches:
\begin{enumerate}[leftmargin=*]
	\setlength\itemsep{0em}
\item \textbf{Unsupervised representation learning:} \tool{} learns graph embeddings in a completely unsupervised manner i.e., class labels of graphs are not required for learning their embeddings. This allows us to readily use \tool{} embeddings in a plethora of applications where labeled data is difficult to obtain.

\item \textbf{Task-agnostic embeddings:} Since \tool{} does not leverage on any task-specific information (e.g., class labels) for its representation learning process, the embeddings it provides are generic.  This allows us to use these embeddings across all analytics tasks involving whole graphs. In fact, \tool{} embeddings could be used to seed supervised representation learning approaches such as \cite{patchy}.

\item \textbf{Data-driven embeddings:} Unlike graph kernels, \tool{} learns graph embeddings from a large corpus of graph data. This enables \tool{} to circumvent the aforementioned disadvantages of handcrafted feature based embedding approaches.

\item \textbf{Captures structural equivalence:} Unlike approaches such as \sv{} \cite{sub2vec} which sample linear substructures (e.g., fixed length random walks) in a graph and learns graph embeddings from them, our framework samples and considers non-linear substructures, namely, rooted subgraphs for learning embeddings. Considering such non-linear substructures are known to preserve \textit{structural equivalence}\footnote{For instance, Weisfeiler-Lehman kernel uses non-linear substructures for computing kernels across graphs and is demonstrated to outperform linear substructure kernels such as random walk kernel and shortest path kernel in many tasks \cite{wlk,dgk}.} and hence this ensures \tool's representation learning process yields similar embeddings for structurally similar graphs.

\end{enumerate}

\textbf{Experiments.} We determine \tool's accuracy and efficiency in both supervised and unsupervised learning tasks with several benchmark and large real-world graph datasets. Also, we perform comparative analysis against several state-of-the-art substructure (e.g., node) representation learning approaches and graph kernels.
Our experiments reveal that \tool{} achieves significant improvements in classification and clustering accuracies over substructure embedding methods and are highly competitive to \soa{} kernels. Specifically, on two real-world program analysis tasks, namely, malware detection and malware familial clustering, \tool{} outperforms state-of-the-art substructure embedding approaches by more than 17\% and 39\%, respectively.

\textbf{Contributions.} We make the following contributions:
\begin{itemize}[leftmargin=*]
	\setlength\itemsep{0em}

\item We propose \tool, an unsupervised representation learning technique to learn distributed representations of arbitrary sized graphs.


\item Through our large-scale experiments on several benchmark and real-world datasets, we demonstrate that \tool{} could significantly outperform substructure representation learning algorithms and highly competitive to \soa{} graph kernels on graph classification and clustering tasks.

\item We make an efficient implementation of \tool{} and the embeddings of all the datasets used in this work publicly available at \cite{ourweb}.
\end{itemize}

The remainder of the paper is organized as follows: In \S \ref{sec:ps} the problem of learning graph embeddings is formally defined. In \S \ref{sec:bg}, preliminaries on word and document representation learning approaches that \tool{} relies on are presented. The proposed method and its evaluation results and discussions are presented in \S \ref{sec:meth} and \S \ref{sec:eval}, respectively. Conclusions are discussed in \S \ref{sec:conc}.


\section{Problem Statement}
\label{sec:ps}
\textit{Given a set of graphs $\mathbb{G} = \{G_1, G_2,...\}$ and a positive integer $\delta$ (i.e., expected embedding size), we intend to learn $ \delta $-dimensional distributed representations for every graph $ G_i \in \mathbb{G} $. The matrix of representations of all graphs is denoted as $ \Phi \in \mathbb{R}^{|\mathbb{G}| \times \delta} $}.

More specifically, let $G = (N, E, \lambda)$, represent a graph, where $N$ is a set of nodes and $E \subseteq (N \times N) $ be a set of edges.
Graph $ G $ is labeled if there exists a function $ \lambda $ such that $\lambda: N \rightarrow \ell$, which assigns a unique label from alphabet $ \ell $ to every node $ n \in N $. Otherwise, $G$ is considered as unlabeled\footnote{Since our procedure requires node labels, in the case of unlabeled graphs, we follow the procedure mentioned in \cite{wlk} and label nodes with their degree.}. Additionally, the edges may also be labeled in which case we also have an edge labeling function, $\eta : E \rightarrow \mathfrak{e} $. 

Given $ G = (N, E, \lambda) $ and $ sg = (N_{sg}, E_{sg}, \lambda_{sg}) $, $ sg $ is a subgraph of $ G $ iff there exists an injective mapping $ \mu: N_{sg} \rightarrow N $ such that $ (n_1, n_2) \in E_{sg} $ iff $ (\mu(n_1), \mu(n_2)) \in E $. In this work, by subgraph, we strictly refer to a specific class of subgraphs, namely, rooted subgraphs. In a given graph G, a rooted subgraph of degree $d$ around node $n \in N$ encompasses all the nodes (and corresponding edges) that are reachable in $d$ hops from $n$.

\section {Background: skipgram word \& document embedding models}
\label{sec:bg}

Our goal is to learn the distributed representations of graphs by extending the recently proposed document embedding techniques in NLP for multi-relational data. Hence, in this section, we review the related background in language modeling, word and document embedding techniques.

\begin{figure*}[t]
	\centering
	\includegraphics[height=3.5cm,width=16cm]{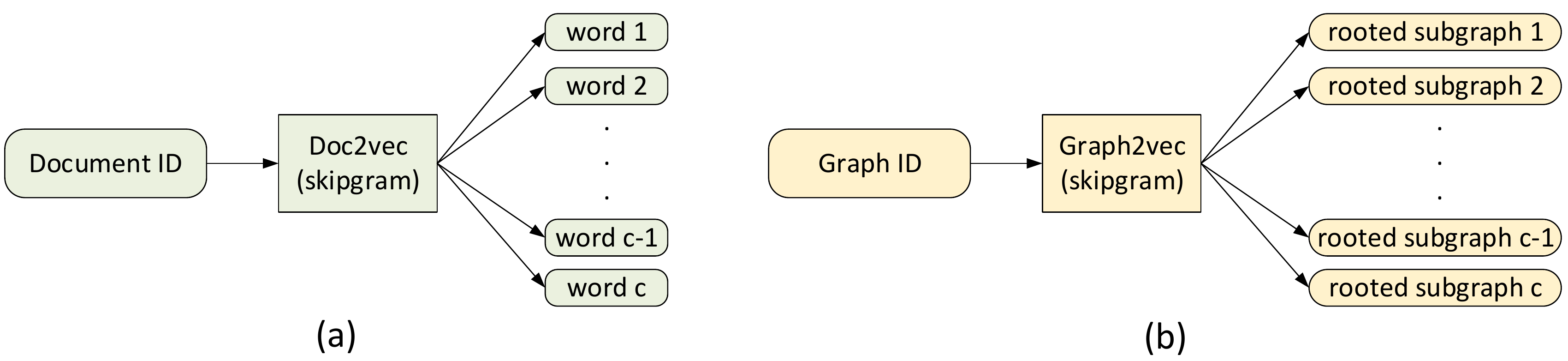}
	\caption{(a) \dv{}'s skipgram model - Given a document \textit{d}, it samples \textit{c} words from \textit{d} and considers them as co-occurring in the same context (i.e., context of the document \textit{d}) uses them to learn \textit{d}'s representation. (b) \tool{} - Given a graph \textit{G}, it samples \textit{c} rooted subgraphs around different nodes that occur in \textit{G} and uses them analogous to \dv{}'s context words and thus learns G's representation.  \label {fig:arch}}
\end{figure*}

\subsection{Skipgram model for learning word embeddings} 
Modern neural embedding methods such as \wv{} \cite{w2v} use a simple and efficient feed forward neural network architecture called "skipgram" to learn distributed representations of words. \wv{} works based on the rationale that \textit{the words appearing in similar contexts tend to have similar meanings and hence should have similar vector representations}.  To learn a target word’s representation, this model exploits the notion of context, where a context is defined as a fixed number of words surrounding the target word. 
To this end, given a sequence of  words $ \{w_1, w_2, ... , w_t, ... , w_T \} $, the target word $w_t$ whose representation has to be learnt and the length of the context window $c$, the objective of skipgram model is to maximize the following log-likelihood:
\begin{equation}
\sum_{t=1}^{T} log \ Pr (w_{t-c},...,w_{t+c}  | w_t)
\end{equation}
where $ w_{t-c},...,w_{t+c} $ are the context of the target word $ w_t $. The probability $ Pr (w_{t-c},...,w_{t+c})$ is computed as
\begin{equation}
\Pi_{-c \le j \le c, j \ne 0} Pr (w_{t+j} | w_t)
\end{equation}
Here, the context words and the target word are assumed to be independent. Furthermore, $ Pr (w_{t+j} | w_t) $ is defined as:
\begin{equation}
\frac {exp(\vec{w_t} \cdot  \vec{w'}_{t+j})} {\sum_{w \in \mathcal{V}} exp(\vec{w_t} \cdot \vec{w})}
\end{equation}

where $ \vec{w} $ and $ \vec{w}{'} $ are the input and output vectors of word $ w $ and $\mathcal{V}$ is the vocabulary of all the words.

\subsection{Negative Sampling}
\label{ss:bg_ns}
The posterior probability in eq. (2) could be learnt in several ways. For instance, a naive approach would be to use a classifier like logistic regression. However, this is prohibitively expensive if the vocabulary $\mathcal{V}$ is very large.

Negative sampling is a simple yet efficient algorithm that helps to alleviate this problem and train the skipgram model. Negative sampling selects a small subset of words at random that are not in the target word's context and updates their embeddings in every iteration instead of considering all words in the vocabulary. Training this way ensures the following: \textit{if a word $ w $ appears in the context of another word $ w' $, then the vector embedding of $ w $ is closer to that of $ w' $ compared to any other randomly chosen word from the vocabulary.}

Once skipgram training converges, semantically similar words are mapped to closer positions in the embedding space revealing that the learned word embeddings preserve semantics. 

\subsection{Neural document embedding models} 
Recently, \dv, a straight forward extension to \wv{} from learning embeddings of words to those of word sequences was proposed by Le and Mikolov \cite{d2v}. \dv{} uses an instance of the skipgram model called paragraph vector-distributed bag of words (PV-DBOW) (interchangeably referred as \dv{} skipgram) which is capable of learning representations of arbitrary length word sequences such as sentences, paragraphs and even whole large documents\footnote{To be precise, there are two versions of \dv{}, namely, PV-distributed memory (PV-DM) and PV-DBOW. PV-DM is not an instance of skipgram model. It learns document embeddings by combining and optimizing them with those of words that occur within a fixed length context window (similar to \wv) in the corresponding documents. Evidently, PV-DM is not related to our proposed technique and hence we refrain from discussing it further.}. More specifically, given a set of documents $D = \{d_1,d_2,...d_N\}$ and a sequence of words $ c (d_i) = \{ w_1, w_2, ..., w_{l_i} \}$ sampled from document $d_i \in D$, \dv{} skipgram learns a $\delta$ dimensional embeddings of the document $d_i \in D$ and each word $w_j$ sampled from $c(d_i)$ i.e., $\vec d_i \in R^\delta$ and $\vec w_j \in R^\delta$, respectively.
The model works by considering a word $w_j \in c(d_i)$  to be occurring in the context of document $d_i$ and tries to maximize the following log likelihood: 

\begin{equation}
\sum_{j=1}^{l_i} log \ Pr (w_j  | d_i)
\end{equation}
where, the probability  $ Pr (w_j  | d)$ is defined as,
\begin{equation}
\frac {exp(\vec{d} \cdot \vec{w_{j}})} {\sum_{w \in \mathcal{V}} exp(\vec{d} \cdot \vec{w})}
\end{equation}
where $\mathcal{V}$ is the vocabulary of all the words across all documents in $D$.
Understandably, eq. (4) could be trained efficiently using negative sampling. 

In \tool, we consider graphs analogical to documents that are composed of rooted subgraphs which, in turn, are analogical words from a special language and extend document embedding models to learn graph embeddings.

\section{Method: Learning Graph Representations}
\label{sec:meth}

In this section we discuss the intuition (\S \ref{ss:intuition}), overview (\S \ref{ss:ov}) and main components of our \tool{} algorithm (\S \ref{ss:algo}) in detail and explain how it learns embeddings of arbitrary sized graphs in an unsupervised manner.

\subsection{Intuition}
\label{ss:intuition}
With the background on word and document embeddings presented in the previous section, an important intuition we extend in \tool{} is to view an entire graph as a document and the rooted subgraphs (that encompass a neighborhood of certain degree) around every node in the graph as words that compose the document. In other words, different subgraphs compose graphs in a similar way that different words compose sentences/documents when used together. 

At this juncture, it is important to note that other substructures such as nodes, walks and paths could also be considered as atomic entities that compose a graph, instead of rooted subgraphs. However, there are two reasons that make rooted subgraphs more amenable for learning graph embeddings:\\
\textbf{1. Higher order substructure.} Compared to simpler lower order substructures such as nodes, rooted subgraphs encompass higher order neighborhoods which offers a richer representation of composition of the graphs. Hence, the embeddings learnt through sampling such higher order substructures would reflect the compositions of the graphs better.\\
\textbf{2. Non-linear substructure.} Compared to linear substructures such as walks and paths, rooted subgraphs capture the inherent non-linearity in the graphs  better. This fact is evident while considering the graph kernels, as well.  For instance, Weisfeiler-Lehman (WL) kernel which are based on non-linear substructures offer significantly better performance on many tasks than the linear substructure based kernels such as random walk and shortest path kernels \cite{wlk,dgk}.

Through establishing the above mentioned analogy of documents and words to graphs and rooted subgraphs, respectively, one can utilize document embedding models to learn graph embeddings. The main expectation here is that structurally similar graphs will be close to each other in the embedding space. In this sense, similar to the Deep Graph Kernels \cite{dgk}, \tool's embeddings provide means to arrive a data-driven graph kernel.

\subsection{Algorithm overview}
\label{ss:ov}
Similar to the document convention, the only required input is a corpus of graphs for
\tool{} to learn their representations. Given a dataset of
graphs, \tool{} considers the set of all rooted subgraphs (i.e., neighbourhoods) around every node (up to a certain degree) as its vocabulary.  Subsequently,  following the \dv{} skipgram training process, we learn the representations of each graph in the dataset.

To train the skipgram model in the above mentioned fashion we need to extract rooted subgraphs and assign a unique label for all the rooted subgraphs in the vocabulary. To this end, we deploy the WL relabeling strategy (which is also used by the WL kernel).

\subsection{\tool: Algorithm}
\label{ss:algo}
The algorithm consists of two main components; first, a procedure to generate rooted subgraphs around every node in a given graph (\S \ref {sss:subgraph_ext}) and second, the procedure to learn embeddings of the given graphs  (\S \ref{ss:sgns}}).

As presented in Algorithm \ref{algo:g2v} we intend to learn $ \delta $ dimensional embeddings of all the graphs in dataset $ \mathbb{G} $ in $ \mathfrak{e} $ epochs. 
We begin by randomly initializing the embeddings for all graphs in the dataset   (line 2). Subsequently, we proceed with extracting rooted subgraphs around every node in each of the graphs (line 8)  and iteratively learn (i.e., refine) the corresponding graph's embedding in several epochs (lines 3 to 10). These steps represent the core of our approach and are explained in detail in the two following subsections. 

\begin{algorithm}[t]
	\scriptsize
	\caption{ \textsc{graph2vec} ($ \mathbb{G}, D, \delta, \mathfrak{e}, \alpha$) \label{algo:g2v}}
	\SetKwInOut{Input}{input}
	\SetKwInOut{Output}{output}
	\Input{$\mathbb{G} = \{G_1, G_2,...,G_n\}$: Set of graphs such that each graph $ G_i = (N_i, E_i, \lambda_i) $ for which embeddings have to be learnt \newline
		$ D $: Maximum degree of rooted subgraphs to be considered for learning embeddings. This will produce a vocabulary of subgraphs, $ SG_{vocab} = \{sg_1, sg_2, ...\} $ from all the graphs in $ \mathbb{G} $\newline
		$\delta$: number of dimensions (embedding size)\newline
		$ \mathfrak{e} $: number of epochs\newline
		$\alpha$: Learning rate}
	\Output{Matrix of vector representations of graphs $ \Phi \in \mathbb{R}^{|\mathbb{G}| \times \delta}$}
	
	\small
	\Begin{
		Initialization: Sample $ \Phi $ from $ R^{|\mathbb{G}| \times \delta} $\\		
		\For{$ e = 1$ to $\mathfrak{e} $}{
			$ \mathfrak{G} $ = \textsc{Shuffle} ($ \mathbb{G} $)\\
			\For {\textbf{each} $G_i \in \mathfrak{G}$}{
				\For {\textbf{each} $ n \in N_i $}{
					\For {$ d = 0 $ to $ D $}{
						$ sg^{(d)}_{n} := $ \textsc{GetWLSubgraph}($n, G_i, d $)\\
						$ J (\Phi) $ = $ - $log Pr ($ sg^{(d)}_n | \Phi(G) $)\\
						$ \Phi = \Phi - \alpha  \frac{\partial J}{\partial \Phi} $
					}
				}
			}
		}
		\textbf{return} $\Phi$
	}	
\end{algorithm}

\begin{algorithm}[t]
	\scriptsize
	\caption{\textsc{GetWLSubgraph} $ (n, G, d) $}
	\label{algo:get_rooted_sg}
	\SetKwInOut{Input}{input}
	\SetKwInOut{Output}{output}
	\Input{$n $: Node which acts as the root of the subgraph  \newline
		$ G = (N,E,\lambda) $: Graph from which subgraph has to be extracted \newline
		$d$: Degree of neighbours to be considered for extracting subgraph }
	\Output{$sg^{(d)}_n$: Rooted subgraph of degree $ d $ around node $ n $ }
	
	\small
	\Begin{
		$ sg^{(d)}_n $ = \{\} \newline
		\If {$d = 0$} {
			$sg^{(d)}_n := \lambda(n) $ 
		}
		\Else{
			$\mathcal{N}_n := \{n'\ |\ (n,n') \in E\}$\\
			$M^{(d)}_n := \{$\textsc{GetWLSubgraph}$(n',G,d-1)\ |\ n' \in \mathcal{N}_n \}$ \\
			$sg^{(d)}_n := sg^{(d)}_n \cup $  \textsc{GetWLSubgraph} $(n,G,d-1) \oplus sort(M^{(d)}_n)$  
		}
		\textbf{return } $ sg^{(d)}_n $
	}
\end{algorithm}


\subsubsection{Extracting Rooted Subgraphs}
\label{sss:subgraph_ext}
To facilitate learning graph embeddings, a rooted subgraph $ sg_{n}^{(d)} $ around every node $ n $ of graph $ G_i $ is extracted (line 8). This is a fundamentally important task in our approach. To extract these subgraphs, we follow the well-known WL relabeling process \cite{wlk} which lays the basis for the WL kernel  \cite{dgk,wlk}. The subgraph extraction process is explained separately in Algorithm \ref{algo:get_rooted_sg}. The algorithm takes the root node $ n $, graph $ G $ from which the subgraph has to be extracted and degree of the intended subgraph $ d $ as inputs and returns the intended subgraph $ sg_n^{(d)} $. When $ d =0 $, no subgraph needs to be extracted and hence the label of node $ n $ is returned (line 3). For cases where $ d > 0 $, we get all the (breadth-first) neighbours of $ n $ in $ \mathcal{N}_n $ (line 5). Then for each neighbouring node, $ n' $, we get its degree $ d-1 $ subgraph and save the same in list $ M^{(d)}_n $ (line 6). Finally, we get the degree $ d-1 $ subgraph around the root node $ n $ and concatenate the same with sorted list $ M_n^{(d)} $ to obtain the intended subgraph $ sg_n^{(d)} $ (line 7).  

\subsubsection{Skipgram with Negative Sampling}
\label{ss:sgns}

Given that $sg^{(d)}_{n} \in SG_{vocab}$ and $ |SG_{vocab}| $ is very large, calculating $ Pr (sg^{(d)}_{n} | \Phi(G))$ in line 9 of Algorithm \ref{algo:g2v} is prohibitively expensive. Hence we follow the negative sampling strategy (introduced in \S \ref{ss:bg_ns}) to calculate this  posterior probability.

In our negative sampling phase, for every training cycle of Algorithm \ref{algo:g2v}, given a graph $G_i \in \mathbb{G}$ and a set of rooted subgraphs in its context,  $c(G_i) = c = \{sg_1, sg_2, ... \}$, we select a set of fixed number of randomly chosen subgraphs as negative samples $c'= \{sgn_1, sgn_2, ... sgn_k \}$  such that $c' \subset SG_{vocab}, k << |SG_{vocab}|$ and $c \cap c' = \{\}$. 
Intuitively, negative samples ($c'$) is a set of $k$  subgraphs, each of which is not present in the graph whose embedding has to be learnt ($G_i$), but in the vocabulary of subgraphs. The number of negative samples ($k$) is a hyper-parameter that could be empirically tuned. For efficient training, for every graph $G_i \in \mathbb{G}$, first, the target-context pairs ($G_i$, $c$) are trained and the embeddings of the corresponding subgraphs are updated. Subsequently, we update only the embeddings of the negative samples $c'$, instead of the whole vocabulary.

Given a pair of graphs $G_i$ and $G_j$, this training makes their embeddings $\Phi(G_i)$ and $\Phi(G_j)$ closer if they are composed of similar rooted subgraphs (i.e., $c(G_i)$ is similar to $c(G_j)$) and at the same time  distances them from the embeddings of all the graphs which are not composed of similar subgraphs.

\subsubsection{Optimization}
\label{sss:opt}
Stochastic gradient descent (SGD) optimizer is used to optimize the parameters in line 9 and 10 of Algorithm \ref{algo:g2v}. The derivatives are estimated using the back-propagation algorithm. The learning rate $ \alpha $ is empirically tuned.

\subsection{Use cases}
Once the embeddings of graphs are computed using \tool, they could  be used for a variety of downstream graph analytics tasks. The prominent ones are reviewed below.

\textbf{Graph Classification.} Given $ \mathbb{G} $, a set of graphs and $Y$, the set of corresponding class labels, graph classification is the task where we learn a model $ \mathcal{H} $ such that $ \mathcal{H}: \mathbb{G} \rightarrow Y $. To this end, one could obtain the embeddings of all the graphs in $\mathbb{G}$ and feed them to general purpose classifiers such as  RFs, Nueral Networks and SVMs to perform classification.
At this juncture, it is important to note that other graph embedding procedure such as graph kernels and substructure embeddings do not offer this flexibility. More specifically, in the case of such methods, the kernel matrices computed using them\footnote{see \cite{dgk} for the explanations on obtaining kernel matrix with substructure embedding approaches} could be used only in conjunction with kernel classifiers (e.g., SVMs) and general purpose classifiers could not be used.

\textbf{Graph Clustering.} Given $ \mathbb{G} $, in graph clustering, the goal is to group similar graphs together. \tool's embeddings could be used along with general purpose clustering algorithms such as K-means and relational clustering algorithms such as Affinity Propagation (AP) \cite{ap} to achieve this. Again, due to the aforementioned limitations of graph kernels and substructure embeddings, they could be used only with relational clustering algorithms.


\section{Evaluation}
\label{sec:eval}

We evaluate \tool's accuracy and efficiency both in graph classification and clustering tasks. Besides experimenting with benchmark datasets, we also evaluate our approach on two real-world  graph analytics tasks from the field of program analysis, namely, malware detection and  malware familial clustering on large  malware datasets. 

\textbf{Research Questions.} Specifically, we intend to address the following research questions: (1) How does \tool{} compare to \soa{} substructure representation learning approaches and graph kernels for graph classification tasks in terms of accuracy and efficiency on benchmark datasets, (2) How does \tool{} compare to the aforementioned \soa{} approaches on a real-world graph classification task, namely, malware detection detection, and (3) How does \tool{} compare to the aforementioned \soa{} approaches on a real-world graph clustering task, namely, malware familial clustering.

\textbf{Comparative Analysis.}
The proposed approach is compared with two representation learning techniques, namely, \nv{} \cite{n2v} and \sv{} \cite{sub2vec} and two graph kernel techniques, namely, WL kernel \cite{wlk} and Deep WL kernel \cite{dgk}. \nv{} is a neural embedding framework which learns feature representation of individual nodes in graphs. In our experiments, to obtain embeddings of entire graphs using \nv{}, we average those of all the nodes in the graph. \sv{} \cite{sub2vec} is a framework that learns representations of any arbitrary subgraphs. Therefore, obtaining representation of whole graphs using \sv{} is a straightforward procedure. 
WL kernel \cite{wlk} is handcrafted feature based kernel that decomposes graphs into rooted subgraphs and computes the kernel values based on them. Besides kernel values, it also yields explicit vector representations of graphs. Deep WL kernel \cite{dgk} is a representation learning variant of WL kernel which learns embeddings of rooted subgraphs such that similar subgraphs have similar embeddings. Thus, the kernel values obtained using subgraph embeddings would be unaffected by the limitations of handcrafted features such as diagonal dominance.

\textbf{Evaluation Setup.}
All the experiments were conducted on a server with 36 CPU cores (Intel E5-2699 2.30GHz processor) and 200 GB RAM running Ubuntu 14.04.

\begin{table}[t]
	\centering
	\setlength\tabcolsep{4pt}
	\scriptsize
	\caption{Benchmark dataset statistics}
	\label{tab:bm_ds}
	\begin{tabular}{|c|c|c|c|}
		\hline
		\textbf{Dataset}  & \textbf{\# samples} & \textbf{\begin{tabular}[c]{@{}l@{}}\# nodes\\   \ \ (avg.)\end{tabular}} & \textbf{\begin{tabular}[c]{@{}l@{}}\# distinct \\ node labels\end{tabular}} \\ \hline \hline
		\textbf{MUTAG}    & 188                 & 17.9                     & 7  \\ 
		\textbf{PTC}      & 344                 & 25.5                     & 19\\ 
		\textbf{PROTEINS} & 1113                & 39.1                     & 3 \\ 
		\textbf{NCI1}     & 4110                & 29.8                     & 37 \\ 
		\textbf{NCI109}   & 4127                & 29.6                     & 38  \\ \hline
	\end{tabular}
\end{table}

\subsection{Graph Classification with Benchmark Datasets}
\label{ss:cl_bmd}
\begin{table*}[t]
	\setlength\tabcolsep{8 pt}
	\centering
	\scriptsize
	\caption{Average Accuracy ($\pm$ std dev.) for \tool{} and state-of-the-art graph kernels on benchmark graph classification datasets}
	\label{tab:res_bm}
	\begin{tabular}{|l|l|l|l|l|l|}
		\hline
		\textbf{Dataset}                     & \textbf{MUTAG} & \textbf{PTC} & \textbf{PROTEINS} & \textbf{NCI1} & \textbf{NCI109} \\ \hline \hline
		
		\textbf{\nv{} \cite{n2v}}                          & 72.63 $\pm$ 10.20   & 58.85 $\pm$ 8.00 & 57.49 $\pm$ 3.57      & 54.89 $\pm$ 1.61  & 52.68 $\pm$ 1.56    \\
		
		\textbf{\sv{} \cite{sub2vec}}                          & 61.05 $\pm$ 15.79   & 59.99 $\pm$ 6.38 & 53.03 $\pm$ 5.55       & 52.84 $\pm$ 1.47  & 50.67 $\pm$ 1.50    \\ 
		
		\textbf{WL kernel \cite{wlk}}                          & 80.63 $\pm$ 3.07   & 56.91 $\pm$ 2.79 & 72.92 $\pm$ 0.56     & 80.01 $\pm$ 0.50  & 80.12 $\pm$ 0.34   \\ 
		
		\textbf{Deep WL kernel \cite{dgk}}                     & 82.95  $\pm$ 1.96    & 59.04 $\pm$ 1.09  & \textbf{73.30} $\pm$ 0.82    & \textbf{80.31} $\pm$ 0.46  & \textbf{80.32} $\pm$ 0.33  \\ 
		\textbf{\tool{}}                & \textbf{83.15} $\pm$ 9.25  & \textbf{60.17} $\pm$ 6.86  & \textbf{73.30} $\pm$ 2.05  & 73.22 $\pm$ 1.81 & 74.26 $\pm$ 1.47 \\ \hline
	\end{tabular}
\end{table*}

\textbf{Datasets.}
Five benchmark graph classification datasets namely MUTAG, PTC, PROTEINS, NCI1 and NCI109 are used in this experiment. 
These datasets belong to chemo- and bio-informatics domains and the specifications of the datasets used are given in Table \ref{tab:bm_ds}. MUTAG is a data set of 188 chemical compounds labeled according to whether or not they have a mutagenic effect on a specific bacteria. PTC dataset comprises of 344 compounds and their classes indicate carcinogenicity on rats. PROTEINS is a collection of graphs whose nodes represent secondary structure elements and edges indicate neighborhood in the amino-acid sequence or in 3D space. NCI1 and NCI109 datasets consist of 4,110 and 4,127 graphs respectively, representing two balanced subsets of datasets of chemical compounds screened for activity against non-small cell lung cancer and ovarian cancer cell lines, respectively. 

\textbf{Experiment \& Configurations.} In this experiment, for each of the datasets, we train a SVM classifier with 90\% of the samples  chosen at random and evaluate its performance on the test set  of remaining 10\% samples.  
The hyper-parameters of the classifiers are tuned based on 5-fold cross validation on the training set.
For all the representation learning methods, we used a common embedding dimensions of 1024, which was arrived empirically\footnote{Embedding dimensions of \{16, 32, 64,..., 4096\} were experimented with and 1024 was found produce best results predominantly with reasonable efficiency, across all datasets}.

\begin{figure}[t]
	\centering
	\includegraphics[height=5cm,width=8.5cm]{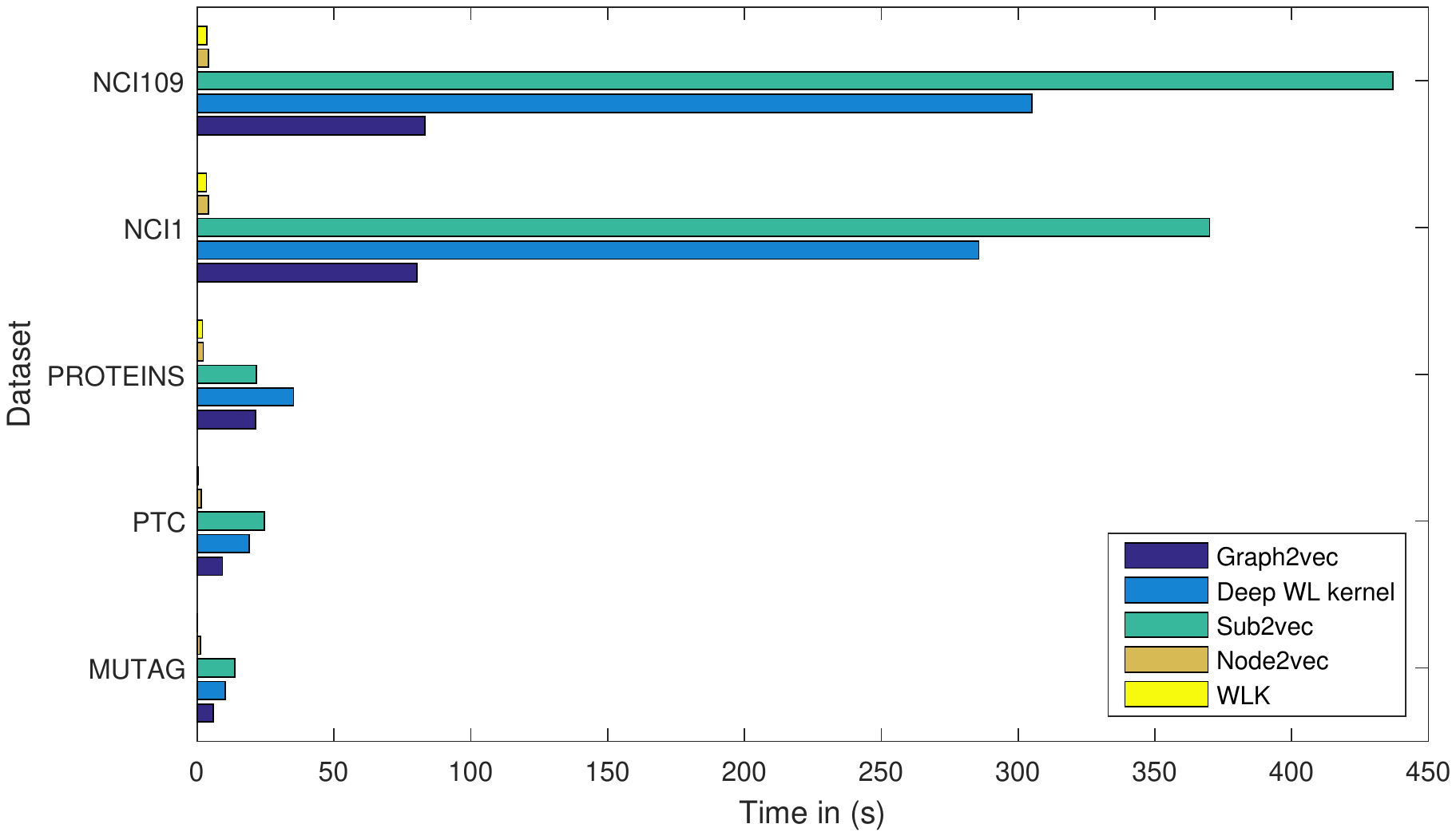}
	\caption{Pre-training durations of graph embedding techniques \label {fig:time}}
\end{figure}

\textbf{Evaluation Metrics.} The experiment is repeated 5 times and the average accuracy is used to determine the effectiveness of classification. Efficiency is determined in terms of time consumed for building graph embeddings (\textit{aka} pre-training duration). The training and testing durations are not reported as they are not directly related to the proposed method.

\subsubsection{Results and Discussion}
\textbf{Accuracy. }The results obtained by the \tool{} on benchmark datasets are summarized in Table \ref{tab:res_bm}. From the results, it is evident that the proposed approach outperforms other representation learning and kernel based techniques on 3 datasets (MUTAG,  PTC and PROTEINS) and has comparable accuracy on the remaining  2 datasets (NCI1 and NCI109). The following inferences are made from the table.

\begin{itemize}
	[leftmargin=*]
	\setlength\itemsep{0em}
	
	\item \nv{} being a lower order substructure embedding technique, it could only model local similarity within a confined neighborhood and fails to learn global structural similarities that helps to classify similar graphs together. This is especially evident from the results on larger datasets, PROTEINS, NCI1 and NCI109 where \nv's accuracy is just above 50\% (i.e., only marginally better than random classification). In general, from these results, one could conclude that while the substructure embeddings techniques excel in substructure based analytics tasks (see \cite{n2v} for \nv's node classification and link prediction performances), extending them directly for tasks involving whole graphs yields sub-par accuracies.
	
	\item \sv{} performs predominantly poorly across all datasets. This is mainly because of the fact that its strategy to sample  graph substructures and learn their embeddings is particularly ill-suited for obtaining embedding of large graphs. That is, \sv{} samples only one random walk (of fixed length) from the given graph and subsequently learns its representations using fixed length linear context skipgrams (with several iterations) over the sampled walk. This prevents \sv{} from learning meaningful representations of an entire graph, as sampling only random walk may not be enough to cover all the neighborhoods in the graph. This ultimately prevents the method from appropriately modeling the structural similarities across graphs which reflects in its poor performance. Also, this inference is reinforced by the fact that \sv{} accuracies decrease with the increase in the size of the graphs (see the difference in accuracies for MUTAG and NCI109 datasets).

	\item WL kernel, being a technique particularly designed to cater tasks such as graph classification, consistently provides good results on all datasets. Deep WL Kernel performs better than WL kernel on all datasets, as it addresses the limitations of the latter kernel's handcrafted features and achieves better generalization. 
	
	\item Finally, \tool's structure-preserving, data-driven embedding  which appropriately models both local and global similarities among  graphs, consistently yields good results on all datasets. In particular, it outperforms all the \soa{} methods in MUTAG, PTC and PROTEINS dataset and obtains slightly lesser accuracies on NCI1 and NCI109 datasets than the kernels.	
\end{itemize}

\textbf{Efficiency.} A pre-training phase to compute vectors of substructures and graphs is required for all the aforementioned methods except WL kernel. On the other hand, WL kernel requires a phase to extract rooted subgraph features and build handcrafted embeddings.
In the case of former approaches, pre-training is the crucial step that helps in capturing latent substructure similarities in graphs and thus aids them to outperform handcrafted feature techniques. Therefore, it is important to study the cost of pre-training. The results of pre-training/feature extraction durations for all the methods under study are shown in Figure \ref{fig:time}. 
 
Understandably, WL kernel is the most scalable method for obtaining graph embeddings as it does not involve learning representations.
\nv{} learns embeddings of lower order entities (i.e., nodes) through confined explorations of neighborhoods around them and hence takes very less time for pretraining. \sv{} learns graph embeddings by sampling linear substructures and running several iterations of skipgram algorithm over them. This results in significantly high pretraining durations. DeepWL kernel learns rooted subgraph embeddings using skipgram. It takes much lesser duration than \sv{} as the latter's length of sampled random walk is much longer than the number of samples rooted subgraphs in the former.  Finally, our approach, which learns embeddings of higher order structures remains less scalable than \nv, but much more scalable than Deep WL kernel and \sv. This is due to the fact that, our approach runs skipgram training only a limited number times (which is equal to the number of rooted subgraphs sampled form the given graph), while the other two approaches run it several times over a fixed length linear context window.  

The efficiency results in our experiments with real-world datasets discussed in subsequent subsections follow the same pattern as the one discussed above. Hence, we refrain from discussing efficiency results here after.

\subsection{Graph Classification with Real-world Dataset}
\label{ss:gclass}

The performances of graph embedding approaches on large real-world datasets may be different from the benchmark ones as they are more complex.
Furthermore, benchmark datasets used in \S \ref{ss:cl_bmd} are too small for the data-driven embedding approaches to reap considerable leverage by exploiting the volume of data over the handcrafted approaches. 
Therefore, it is highly essential to evaluate the performance of the proposed method on large real-world datasets to showcase its true potentials.

\textbf{Experiment \& Configurations.} To this end, we consider a large-scale Android malware detection problem where we are given a dataset of API Dependency Graphs (ADGs) of malicious and benign Android apps, and the task is to  represent each of them as vectors and train ML classifiers to identify malicious ones. 
The datasets statistics are presented in Table \ref{tab:real_ds}. Evidently, these ADGs are much larger than the benchmark graphs. The dataset comprises of 10,560 ADGs, each of which contain more than 2,600 nodes (i.e., instructions), 920 edges (i.e., control flows among instructions) and 4200 unique node labels (i.e., APIs invoked in instructions) on average. The training set comprises of 70\% of samples chosen at random and the remaining 30\% samples are used as test set to evaluate the models. The experiment is repeated 5 times and the results are averaged.

\begin{table}[t]
	\centering
	\setlength\tabcolsep{1pt}
	\scriptsize
	\caption{Large real-world datasets used in graph classifications and clustering tasks}
	\label{tab:real_ds}
	\begin{tabular}{|c|c|c|c|c|c|c|}
		\hline
		\textbf{Dataset} & \textbf{source} & \textbf{\begin{tabular}[c]{@{}c@{}}\# \\ samples\end{tabular}} & \textbf{\begin{tabular}[c]{@{}c@{}}\#\\ classes\end{tabular}} & \textbf{\begin{tabular}[c]{@{}c@{}}\# nodes\\ (avg.)\end{tabular}} & \textbf{\begin{tabular}[c]{@{}c@{}}\# edges\\ (avg.)\end{tabular}} & \textbf{\begin{tabular}[c]{@{}c@{}}\# distinct\\ node labels\end{tabular}} \\ \hline \hline
		Classification & \cite{drebin,gp} & 10,560 & 2 & 2637.12 & 927.31 & 4271 \\ 
		Clustering & \cite{amd} & 24,650 & 71 & 1071.33 & 544.83 & 3828 \\ \hline
	\end{tabular}
\end{table}

\textbf{Evaluation Metrics.} The same evaluation metrics that are used in experiments with benchmark datasets (see \S \ref{ss:cl_bmd}) are used here as well.

\begin{table}[t]
	\setlength\tabcolsep{5pt}
	\centering
	\scriptsize
	\caption{Malware Detection - Results}
	\label{tab:maldetect_res}
	\begin{tabular}{|l|c|}
		\hline
		\textbf{Method} & \textbf{Accuracy (avg. $\pm$ std.)}  \\ \hline \hline
		
		\nv{} \cite{n2v} & 81.25  $\pm$ (1.04)\\
		\sv{} \cite{sub2vec} & 76.83  $\pm$  (2.83)\\
		WL kernel \cite{wlk} & 97.12 $\pm$ (0.44) \\
		Deep WL kernel \cite{dgk} & 98.16  $\pm$ (0.20)\\
		\tool & \textbf{99.03}  $\pm$ \textbf{(0.17)}\\ \hline
	\end{tabular}
\end{table}

\subsubsection{Results \& Discussion}

The malware detection results of the proposed and compared \soa{} approaches are presented in Table \ref{tab:maldetect_res}. From the results obtained, the following inference is drawn:
\begin{itemize}
	[leftmargin=*]
	\setlength\itemsep{0em}
	
	\item Averaging \nv{} embeddings and using \sv{} to obtain graph representations perform poorly in this experiment as well. In particular, the proposed approach outperforms these two techniques by more than 17\% and 22\%, respectively.
	
	\item Both WL and Deep WL kernels perform significantly better than the two substructure representation learning approaches. However, \tool{} still outperforms these techniques by 1.91\% and 0.87\%, respectively.
	
	\item Evidently, being data-driven approaches, both \tool{} and Deep WL kernel exhibit excellent performance on this large-scale dataset. Especially in this experiment, the range in which they outperform other techniques under comparison is more pronounced than the experiments with benchmark datasets. Again, the two representation learning approaches, \nv{} and \sv{} perform worse as they are ill-suited for learning embeddings of entire graphs.
	
\end{itemize}

\subsection{Graph Clustering}
\label{ss:gclus}

The goal of this experiment is to demonstrate the efficacy of \tool's embedding in a downstream analytics task where we do not have class labels for graphs. This task would be most appropriate for evaluating and comparing the methods that do not leverage on any task-specific information in the process of learning representations.

\textbf{Experiment \& Configurations.} In this experiment, we are given with ADGs of malware samples and the name of families\footnote{Samples belonging to same families perform similar malicious activities} to which they belong and the task is to group samples belonging to the same family into the same cluster. To this end, we consider the AMD dataset \cite{amd} which comprises of more than 24,000 Android malware apps belonging to 71 families. The statistics of this dataset is presented in Table \ref{tab:real_ds}. 

From this dataset, only the large families that have more than 100 corresponding malware samples are considered for clustering as this helps to mitigate imbalance in the cluster sizes. The embeddings and kernels of ADGs belonging to these families are built and a relational clustering algorithm, namely, AP \cite{ap} is used to cluster them.

\textbf{Evaluation Metric.} In order to quantitatively measure malware familial clustering accuracy, a standard clustering evaluation metric, namely, Adjusted Rand Index (ARI) is used. The ARI values lie in the range [-1, 1]. For the ease of understanding, we report the ARI as a percentage value. A higher ARI means a higher correspondence to the ground-truth data.

\begin{table}[t]
	\setlength\tabcolsep{5pt}
	\centering
	\scriptsize
	\caption{Malware Clustering - Results}
	\label{tab:malclust_res}
	\begin{tabular}{|l|c|}
		\hline
		\textbf{Method} & \textbf{ARI (as \%)}  \\ \hline \hline
		\nv{} \cite{n2v} & 16.39 \\
		\sv{} \cite{sub2vec} & 14.55 \\
		WL kernel \cite{wlk} & 48.93 \\
		Deep WL kernel \cite{dgk} & 50.41 \\
		\tool & \textbf{56.28} \\ \hline
	\end{tabular}
\end{table}

\subsubsection{Results and Discussion}
The results of malware clustering using \tool{} and other state-of-the-art methods are presented in Table \ref{tab:malclust_res}. From the table, the following inferences are drawn:
\begin{itemize}
	[leftmargin=*]
	\setlength\itemsep{0em}
	
	\item At the outset, we observe all the method obtain lesser ARIs in this experiment, as the malware clustering task is inherently more complex than two classification tasks considered previously. 
	
	\item Similar to the classification tasks, both \nv{} and \sv{} perform poorer than the kernels and \tool. This reinforces the  inference that adopting \nv{} and \sv{} for graph embedding will yield subpar results.

	\item Both WL and Deep WL kernel perform much better than the two aforementioned embedding approaches. However, different from the classification tasks, in this task, the former methods outperform the latter methods by more than 2 folds. 
	
	\item In this experiment, \tool{} outperforms all the other compared approaches highly significantly. In particular, it outperforms the substructure embedding techniques by more than 39\% and the kernels by more than 5\%. 
	This reinforces the findings inferred from the classification experiments. 
\end{itemize}

\textbf{Summary.} Summarizing the inferences from all the three experiments, one could see: (1) trivially extending node and subgraph representation learning approaches to build graph embeddings yield sub-par results, and (2) learning graph embedding from data leads to more accurate results than building the same using handcrafted features. Since \tool{} is appropriately designed, it achieves excellent accuracies in graph analytics tasks with reasonably good efficiency.

\section{Conclusions}
\label{sec:conc}
In this paper, we presented \tool, an unsupervised representation learning technique to learn embedding of  graphs of arbitrary sizes.
Through our large-scale experiments involving benchmark graph classification datasets, we demonstrate that graph embeddings learnt by our approach outperform substructure embedding approaches significantly and are comparable to graph kernels. Since \tool{} is a data-driven representation learning approach, its true potentials are revealed when trained on large volumes of graphs. To this end, when evaluated on two real-world applications involving large graph datasets, \tool{} outperforms state-of-the-art graph kernels without compromising efficiency of the overall performance. We make all the code and data used within this work available at: \cite{ourweb}.


\section{Acknowledgments}
\label{sec:ack}
We thank the authors of \cite{n2v}, \cite{dgk} and \cite{wlk} for making the source code of their approaches publicly available. We thank the authors of \cite{sub2vec} for sharing their approach's source code with us.

\end{document}